\documentclass[sigconf]{acmart}
\AtBeginDocument{%
  }
\usepackage{multirow}
\setcopyright{acmlicensed}

\copyrightyear{2026}
\acmYear{2026}
\setcopyright{cc}
\setcctype{by}
\acmConference[MM '26] {Proceedings of the 34th ACM International Conference on Multimedia}{November 10--14, 2026}{Rio de Janeiro, Brazil.}
\acmBooktitle{Proceedings of the 34th ACM International Conference on Multimedia (MM '26), November 10--14, 2026, Rio de Janeiro, Brazil}
\acmISBN{979-8-4007-2213-4/2026/11}
\acmDOI{xx.xxxx/xxxxxxx.xxxxxxx}

\usepackage{pifont}
\usepackage{bbding}
\usepackage{xcolor}
\usepackage{adjustbox}
\usepackage{multirow}
\usepackage{colortbl}
\definecolor{mycolor1_1}{RGB}{229, 244, 237}
\definecolor{mycolor1_2}{RGB}{255, 229, 229}
\definecolor{forestgreen}{rgb}{0.0, 0.5, 0.0}
\definecolor{myblue}{RGB}{208, 64, 64}

\usepackage{framed}
\usepackage{tcolorbox}
\newtcolorbox{commentbox}{
    colback=gray!10,          
    colframe=black,         
    arc=2pt,                  
    boxrule=0.5pt,            
    left=2pt,                 
    right=2pt,                
    top=2pt,                  
    bottom=2pt,               
    boxsep=1pt,               
    fontupper=\itshape,       
}
\begin{document}

\title[Understanding and Overcoming Cross-modal Fusion Bias in Multimodal Anomaly Detection]{Understanding and Overcoming Cross-modal Fusion Bias in Multimodal Anomaly Detection From A Fisher Information Perspective}


\author{Kaifang Long}
\affiliation{%
  \institution{Northeastern University}
  \city{Shenyang}
  \country{China}}
\email{kaifanglong@163.com}

\author{Lianbo Ma}
\authornote{Corresponding author.}
\affiliation{%
  \institution{Northeastern University}
  \city{Shenyang}
  \country{China}
}
\email{malb@swc.neu.edu.cn}

\author{Liming Liu}
\affiliation{%
  \institution{Northeastern University}
  \city{Shenyang}
  \country{China}}
\email{liuliming@stumail.neu.edu.cn}

\author{Guoyang Xie}
\affiliation{%
  \institution{CATL}
  \city{Ningde}
  \country{China}}
\email{guoyang.xie@ieee.org}

\renewcommand{\shortauthors}{Kaifang Long et al.}

\begin{abstract}
Current advancements in Multimodal Anomaly Detection (MAD) are largely driven by enhancing multimodal fusion, particularly through the integration of RGB and Depth data for richer anomaly representation. However, less attention was devoted to analyzing the role of cross-modal fusion bias—a well-known challenge in multimodal learning—in MAD. This gap motivates a key question: can we overcome this bias to break the performance bottleneck of current work? In this paper, we first analyze the impact of cross-modal fusion bias in MAD via the Fisher Information Matrix. Then, grounded in these findings, we propose UCFB, a simple yet effective plug-and-play framework designed to mitigate cross-modal fusion bias in MAD. It achieves this by jointly employing Fisher-information-guided dynamic calibration to adjust modality-specific regularization weights and canonical similarity analysis to improve inter-modal interactions. Extensive experiments on the MVTec 3D-AD and Eyecandies datasets demonstrate that UCFB achieves consistent improvements in single-class,  multi-class, and few-shot settings. 
\end{abstract}



\maketitle

\section{Introduction}
\label{sec:intro}
\textbf{Limitation.} Multimodal Anomaly Detection (MAD) \cite{tao2025g2sf, li2025find} has become a pivotal task in intelligent manufacturing, aiming to leverage both RGB and depth images to pinpoint abnormal regions of industrial products. Compared to unimodal approaches that rely solely on RGB \cite{zhangcostfilter, mao2025beyond, jiangmmad} or point cloud data \cite{DBLP:conf/ijcai/ChengGZ0DW25, zheng2025bridging, ye2025po3ad}, MAD delivers richer appearance and geometric features, thereby revealing subtle flaws that are challenging to recognize with a single modality. Despite this potential, most existing efforts are devoted to developing novel multimodal fusion strategies \cite{li2026evolving}, such as early fusion \cite{chu2023shape, cheng2026comprehensive}, hybrid fusion \cite{bergmann2023anomaly, wang2023multimodal}, or more recent architecture-oriented methods \cite{long2025revisiting}. However, the critical issue of \textbf{cross-modal fusion bias} (arising from one modality suppressing another, causing suboptimal performance) has received little attention and remains an unresolved, underexplored challenge in MAD.

\begin{figure*}[t]
  \centering
  \setlength{\abovecaptionskip}{0.1cm}
      \includegraphics[scale=0.60]{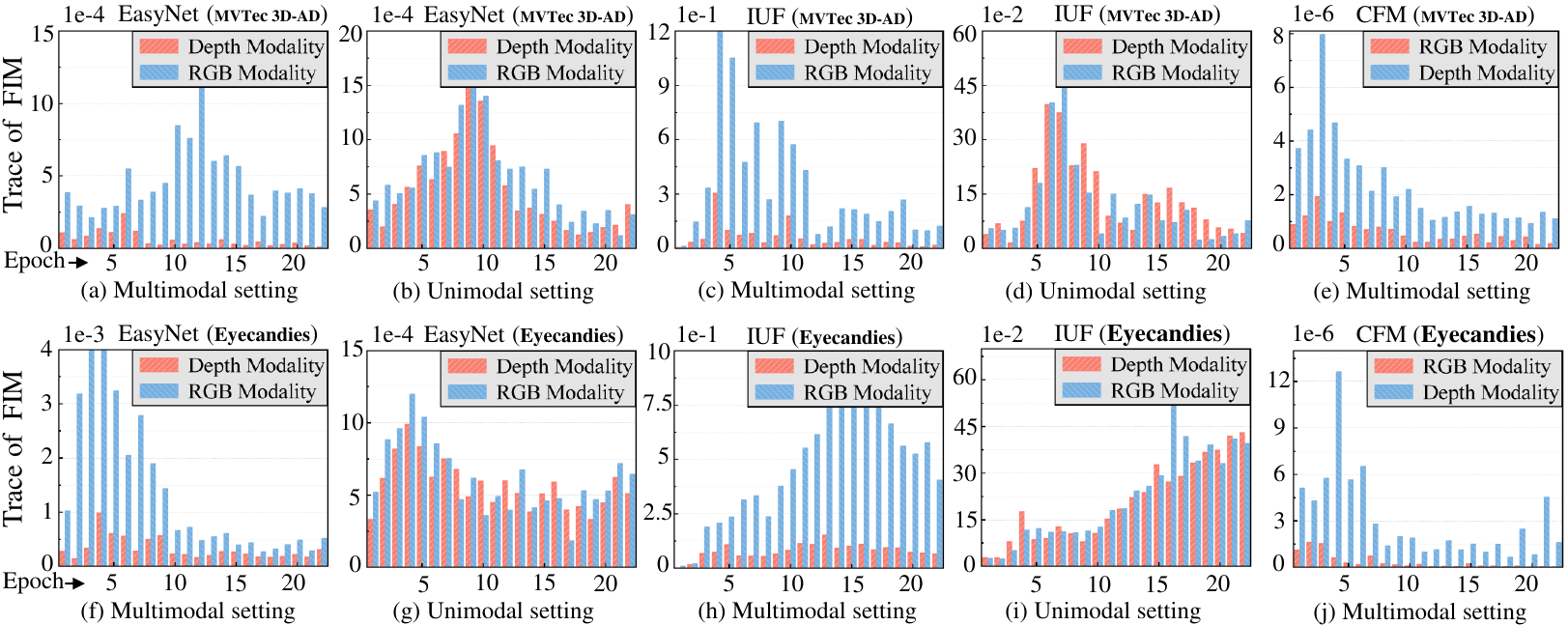}
  \caption{The gap in the trace of FIM between the RGB and depth modalities for three baseline models under unimodal and multimodal settings on the Eyecandies (CandyCane) and MVTec 3D-AD (Bagel) datasets. Due to the limitation of pages, experimental results for other categories are provided in the Supplementary Materials.}
  \label{rgbd_vis_1}
\end{figure*}

\textbf{Motivation.} To advance MAD, this naturally raises an important question: \textbf{\textit{Does cross-modal fusion bias intrinsically limit MAD research?}} In fact, as previewed in Figures \ref{rgbd_vis_1}-\ref{fig1111_1_1_1} and Table \ref{tab1}, we can observe that the cross-modal fusion bias has a clear and substantial negative effect on MAD performance (detailed analysis is provided in \textbf{Section \ref{sec:Validation}}). This motivates us to develop an effective solution to mitigate this bias for MAD. For this purpose, the primary goals of this paper are threefold: (\romannumeral1) \textit{thoroughly revealing the potential impact of cross-modal fusion bias on MAD performance}, (\romannumeral2) \textit{identifying core design principles that can improve MAD}, and (\romannumeral3) \textit{proposing a simple yet effective UCFB framework, which serves as a plug-and-play component for MAD}. We combine theoretical analysis with empirical evaluation and conduct extensive experiments to achieve this goal.

\textbf{Our observations.} Given that CFM \cite{costanzino2024multimodal}, EasyNet \cite{chen2023easynet}, and IUF \cite{tang2024incremental} are well-established representative frameworks in MAD, we initiate this work from them to reveal the issue of cross-modal fusion bias and assess its impact on MAD. Specifically, we first introduce the trace of Fisher Information Matrix (FIM) \cite{DBLP:conf/icml/LiDTR25, huang2025adaptive} to quantify modality-specific information acquisition under both unimodal and multimodal settings, thereby characterizing the potential suppression of one modality by another in MAD. Then, five carefully designed ablative cases, together with visualization, are conducted to systematically reveal the impact of cross-modal fusion bias on the overall learning process and performance in MAD. To ensure the generalizability and reliability of our experimental findings, we evaluate them across three mainstream MAD methods on two datasets. Based on our empirical results, we identify key design principles for mitigating cross-modal fusion bias and derive the following observations:

\ding{182} In the unimodal setting, the baselines exhibited similar information acquisition levels and overall variation trends when trained solely on RGB or depth. (Figure \ref{rgbd_vis_1}) \footnote{For more experimental results and related supplementary materials, please refer to: https://github.com/longkaifang/UCFB}.

\ding{183} In multimodal settings, jointly training RGB and depth often induces modality suppression, where one modality dominates the other; specifically, this manifests as one modality acquiring significantly more information than the other during the early training stages. (Figure \ref{rgbd_vis_1}) 


\ding{184} Interfering with the learning process at early stages will persistently degrade the overall performance of MAD, as the model's ability to integrate multimodal information critically hinges on exposure to task-relevant and consistent signals during this phase. (Case 1-4 in Table \ref{tab1})  

\ding{185} Multimodal fusion performance is highly dependent on the information acquisition of each modality in the early stages; once dominance modality emerges, simply extending the training period cannot compensate for the suppression of one modality by another. (Figure \ref{fig1111_1_1_1}-b, case 5 in Table \ref{tab1})

\ding{186} Identifying a critical learning epoch for mitigating cross-modal fusion bias, namely, the early stage of model training. Targeted adjustment at this stage substantially reduces fusion bias and advances overall MAD performance.

\ding{187} In summary, the suboptimal performance caused by cross-modal fusion bias primarily stems from one modality suppressing another at the early stage; therefore, dynamically calibrating the modality-specific weights at this stage is crucial to mitigate this bias and improve fusion performance.


\textbf{ Contributions.} Based on the above insights, we demonstrate the negative impact of cross-modal fusion bias on MAD and elucidate its underlying mechanism. Then, another question is arisen: \textbf{How can we efficiently mitigate this bias to break the performance bottleneck of current works?} A straightforward and intuitive solution is to slow information acquisition of the dominant modality while accelerating that of the suppressed one during the critical learning epoch. Hence, we propose UCFB, a simple yet effective plug-and-play framework designed to mitigate cross-modal fusion bias in MAD. In UCFB, we first assess whether modality suppression is present. If so, we further identify the dominant modality and employ Fisher-information-guided dynamic calibration to adjust modality-specific regularization weights (i.e., unimodal adaptive adjustment) during the critical learning epoch, while utilizing canonical similarity analysis to enhance cross-modal interactions effectively to boost information acquisition in the suppressed modality. The main contributions can be summarized as: 

\begin{itemize}
    \item To the best of our knowledge, this is the first attempt to thoroughly scrutinize the impact of cross-modal fusion bias on MAD from Fisher information view and, via extensive empirical evidence, identify the key design principles to mitigate this bias and boost performance.
    \item We propose a novel Fisher-information-guided MAD-friendly framework, termed UCFB, which mitigates the issue of cross-modal fusion bias through unimodal adaptive adjustment and canonical similarity analysis, and can be seamlessly integrated into various MAD methods as a plug-in component.  
  \item Extensive theoretical and experimental results demonstrate the effectiveness of UCFB in mitigating cross-modal fusion bias and yielding superior performance across single-class, multi-class, and few-shot settings. 
\end{itemize}

\begin{table*}[t!]
\centering
\caption{Performance of five ablative cases on CFM and EasyNet in terms of I-AUROC, P-AUROC, and AUPRO. Blurring the input image introduces a blur operation to the original data to investigate the effect of modality suppression on fusion performance at various stages.}
\setlength{\tabcolsep}{2.2mm} 
\renewcommand{\arraystretch}{1}
\setlength{\cmidrulewidth}{1.0pt}
\adjustbox{width=1\linewidth}{
\begin{tabular}{clccc c clccc}
\toprule &\multicolumn{10}{|c}{\includegraphics[width=19cm]{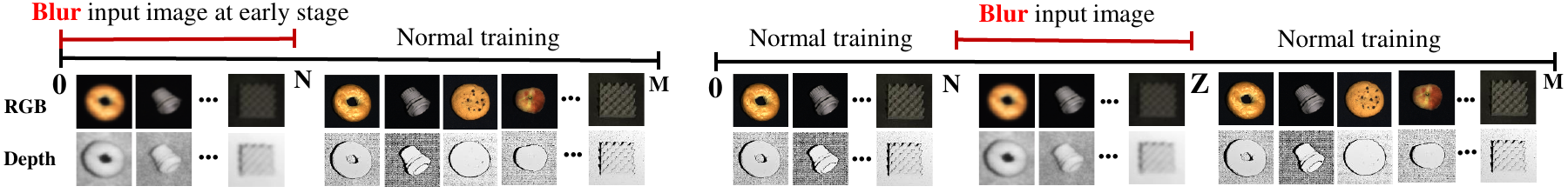}} \\
\midrule
\multirow{5}{*}{Setting}&\multicolumn{10}{|l}{Ablative case 1: model is first trained for N epochs using blurred RGB and original Depth images, then for M epochs with original data.} \\
&\multicolumn{10}{|l}{Ablative case 2: model is first trained for N epochs using blurred Depth and original RGB images, then for M epochs with original data.} \\
&\multicolumn{10}{|l}{Ablative case 3: model is first trained for N epochs using blurred RGB and Depth images, then for M epochs with original data.} \\
&\multicolumn{10}{|l}{Ablative case 4: model is first trained for N epochs using blurred RGB and Depth images, then for M+200 epochs with original data.}\\
&\multicolumn{10}{|l}{Ablative case 5: train for N epochs on original data, then Z epochs with blurred RGB and depth images, finally M epochs on original.}\\

\cmidrule{1-5} \cmidrule{7-11}
\cellcolor{gray!15}&\cellcolor{gray!15} &\multicolumn{3}{c}{\cellcolor{gray!15}MVTec 3D-AD}&  &\cellcolor{gray!15}&\cellcolor{gray!15} &\multicolumn{3}{c}{\cellcolor{gray!15}MVTec 3D-AD}\\
\cellcolor{gray!15}&\multirow{-2}{*}{\cellcolor{gray!15}Methods \& Results}  &\cellcolor{gray!15}I-AUROC  &\cellcolor{gray!15}P-AUROC  &\cellcolor{gray!15}AUPRO \quad\quad   & &\cellcolor{gray!15} &\multirow{-2}{*}{\cellcolor{gray!15}Methods \& Results} &\cellcolor{gray!15}I-AUROC  &\cellcolor{gray!15}P-AUROC  &\cellcolor{gray!15}AUPRO  \\
\cmidrule{1-5} \cmidrule{7-11}     
\multirow{7}{*}{CFM} & \multicolumn{1}{|l|}{Ablative case 1} &\cellcolor{mycolor1_1}88.2 &\cellcolor{mycolor1_1}90.4 &\cellcolor{mycolor1_1}89.5  &&\multirow{7}{*}{EasyNet} & \multicolumn{1}{|l|}{Ablative case 1} &\cellcolor{mycolor1_1}82.5 &\cellcolor{mycolor1_1}80.6 &\cellcolor{mycolor1_1}73.4\\
& \multicolumn{1}{|l|}{Ablative case 2} &\cellcolor{mycolor1_1}86.8 &\cellcolor{mycolor1_1}89.6 &\cellcolor{mycolor1_1}89.1  && & \multicolumn{1}{|l|}{Ablative case 2} &\cellcolor{mycolor1_1}85.4 &\cellcolor{mycolor1_1}83.3 &\cellcolor{mycolor1_1}76.8\\
& \multicolumn{1}{|l|}{Ablative case 3} &\cellcolor{mycolor1_1}78.4 &\cellcolor{mycolor1_1}81.7 &\cellcolor{mycolor1_1}80.0  & && \multicolumn{1}{|l|}{Ablative case 3} &\cellcolor{mycolor1_1}74.8 &\cellcolor{mycolor1_1}75.1 &\cellcolor{mycolor1_1}69.6\\
& \multicolumn{1}{|l|}{Ablative case 4} &\cellcolor{mycolor1_1}79.3 &\cellcolor{mycolor1_1}82.2 &\cellcolor{mycolor1_1}81.2  & && \multicolumn{1}{|l|}{Ablative case 4} &\cellcolor{mycolor1_1}75.0 &\cellcolor{mycolor1_1}74.7 &\cellcolor{mycolor1_1}69.9\\
& \multicolumn{1}{|l|}{Ablative case 5} &\cellcolor{mycolor1_1}94.1 &\cellcolor{mycolor1_1}97.8 &\cellcolor{mycolor1_1}95.7  & && \multicolumn{1}{|l|}{Ablative case 5} &\cellcolor{mycolor1_1}91.8 &\cellcolor{mycolor1_1}90.4 &\cellcolor{mycolor1_1}81.1\\

\cmidrule{2-5} \cmidrule{8-11}
& \multicolumn{1}{|l|}{Baseline} &\cellcolor{mycolor1_1}95.4 &\cellcolor{mycolor1_1}99.3 &\cellcolor{mycolor1_1}97.1  & && \multicolumn{1}{|l|}{Baseline} &\cellcolor{mycolor1_1}92.6 &\cellcolor{mycolor1_1}91.9 &\cellcolor{mycolor1_1}82.1\\
& \multicolumn{1}{|l|}{Baseline+UCFB (Ours)} &\cellcolor{mycolor1_2}96.1 &\cellcolor{mycolor1_2}99.4 &\cellcolor{mycolor1_2}97.5  & && \multicolumn{1}{|l|}{Baseline+UCFB (Ours)} &\cellcolor{mycolor1_2}93.8 &\cellcolor{mycolor1_2}94.6 &\cellcolor{mycolor1_2}84.2\\
\cmidrule{1-5} \cmidrule{7-11}
\end{tabular}
   }
\label{tab1}
\end{table*}
\section{Validation of Cross-modal Fusion Bias Issue}
\label{sec:Validation}
This section describes the experimental setup for the baseline frameworks, followed by a series of empirical analyses to validate the impact of cross-modal fusion bias on MAD.
\subsection{Empirical Study Settings}
\textbf{Benchmark.} CFM \cite{costanzino2024multimodal}, EasyNet \cite{chen2023easynet}, and IUF \cite{tang2024incremental} are selected as the foundational frameworks, given their widespread application in existing efforts. Our primary objective is to systematically analyze and evaluate the impact of cross-modal fusion bias on the overall learning process and performance of MAD using the MVTec 3D-AD and Eyecandies datasets. 

\textbf{Evaluation Metrics.} We evaluate model performance using I-AUROC, P-AUROC, and AUPRO, and quantify the information acquisition of each modality in the model through the trace of FIM, denoted as $Tr(F_{\kappa})$. A larger $Tr(F_{\kappa})$ indicates that the corresponding modality has a stronger impact on the model. $Tr(F_{\kappa})$ is defined as follows:
\vspace{-0.08in}
\begin{equation}
\begin{split}
g_{\varphi^{\kappa}}&(w_{\kappa},x^{\kappa})=\nabla w_{\kappa} \mathcal{L}_{total}(w_{\kappa}),\\
Tr(F_{\kappa})&=\mathbb{E}_{x^{\kappa}\thicksim \chi^{\kappa}}\left[\parallel g_{\psi^{\kappa}}(w_{\kappa},x^{\kappa})\parallel^2\right],\\
\end{split}
  \label{equ:21_111}
\end{equation}
where $x^{\kappa}$ denotes the RGB or depth modality, $\varphi$ represents the modality encoder, $\nabla w_{\kappa}$ is the gradient operator with respect to the parameters of the RGB or depth modality, $\mathcal{L}$ is the overall loss, and 
$||\cdot||^2$ denotes the squared $L2$ norm.

\begin{figure}[t]
\begin{minipage}{0.98\linewidth}
\centerline{\includegraphics[width=\textwidth]{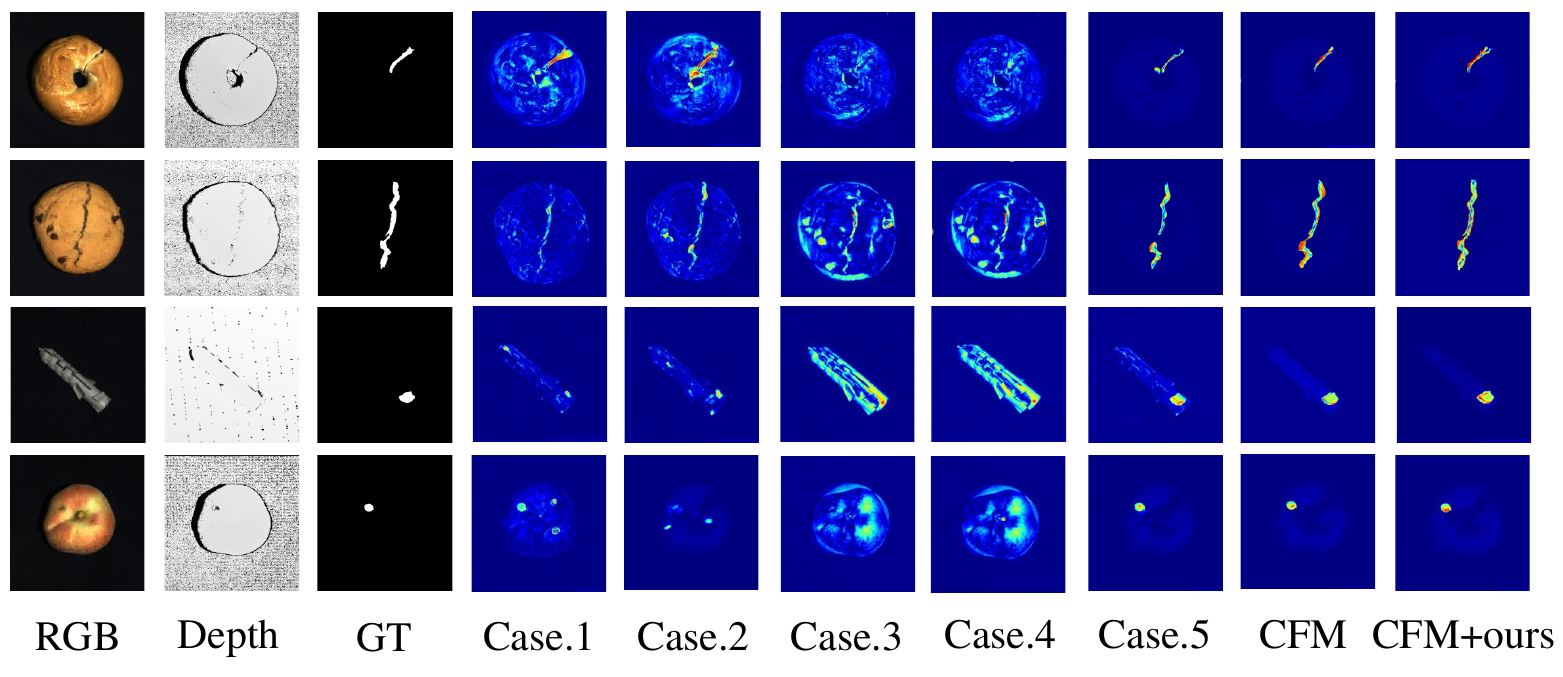}}
	\end{minipage}
    \vspace{2pt}
 \centerline{(a) Visualization of different ablative cases.}
  \begin{minipage}{0.98\linewidth}
		\centerline{\includegraphics[width=\textwidth]{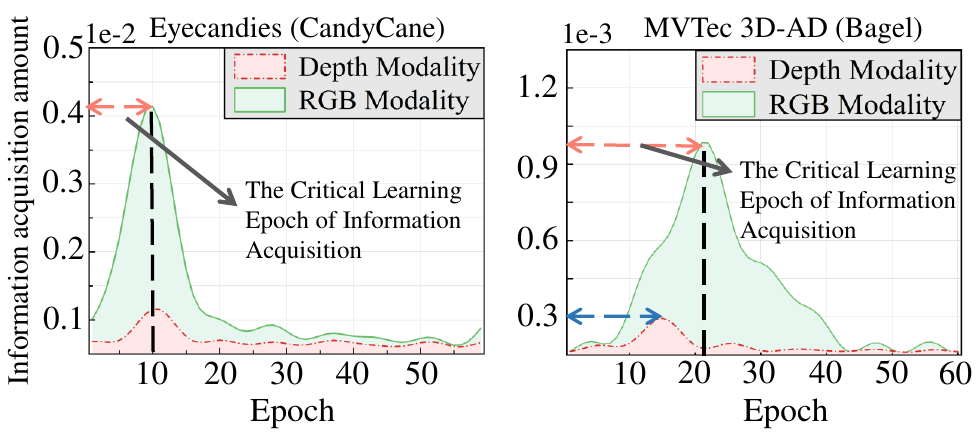}}
	\end{minipage}
 \centerline{(b) Information variation across modalities.}
	\caption{(\textbf{Top}) Visualization of different ablative cases; (\textbf{Bottom}) The variation in information acquisition for each modality during training on EasyNet.}
  \label{fig1111_1_1_1}
\end{figure}
\begin{figure*}[t]
  \centering
  \setlength{\abovecaptionskip}{0.1cm}
      \includegraphics[scale=0.58]{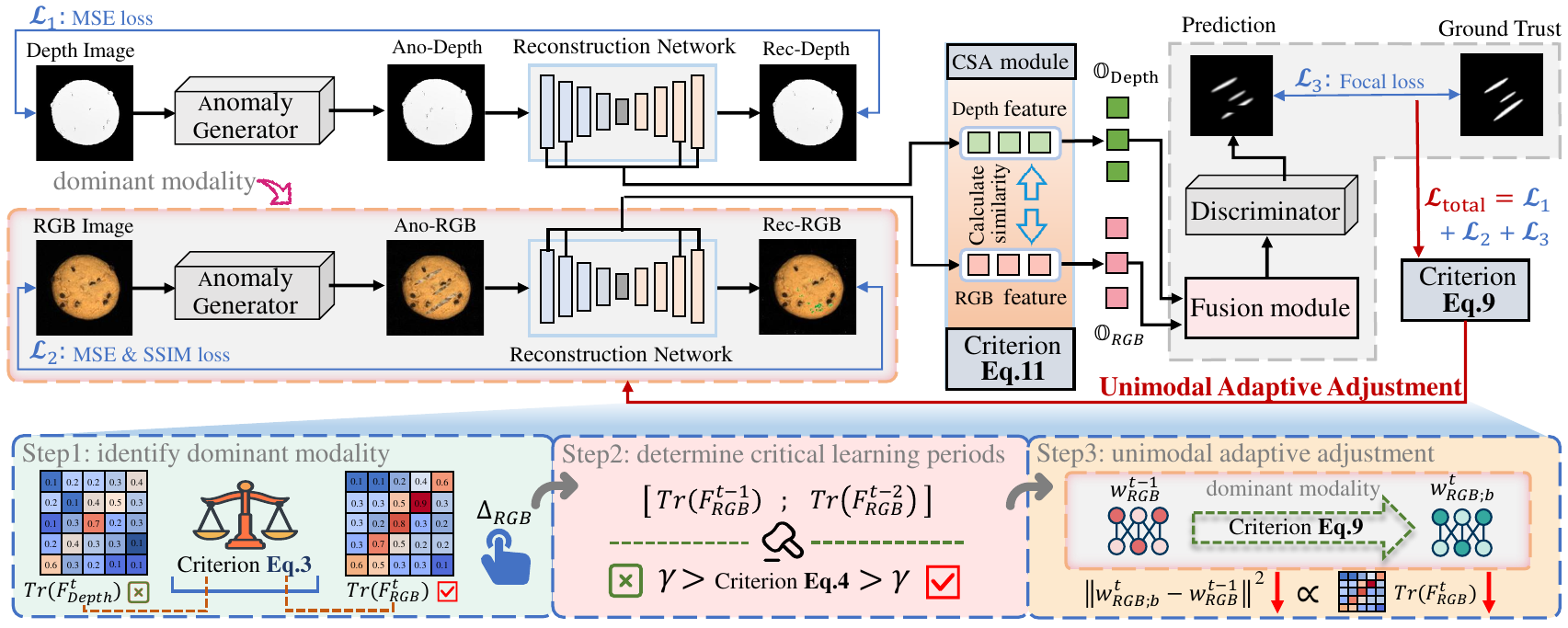}
  \caption{Overview of UCFB. During training, UCFB first identifies the dominant modality (\textbf{Step 1}), checks whether it is in the critical learning period (\textbf{Step 2}), and then applies unimodal adaptive adjustment to slow its information acquisition (\textbf{Step 3}). Simultaneously, the CSA module is employed to strengthen cross-modal interactions, enhancing information acquisition in the suppressed modality.}
  \label{overall}
  \vspace{-0.1in}
\end{figure*}
\textbf{Implementation Setup.} To evaluate the impact of cross-modal bias, we resize the input images to 224×224 for CFM and IUF, and to 256×256 for EasyNet. The models are trained for 50 epochs on CFM and for 1000 epochs on EasyNet and IUF, while all other hyperparameters strictly follow their original settings (see Supplementary Materials for more details). 

\subsection{Empirical Study Results}
\textbf{The Issue of Cross-modal Fusion Bias.} To investigate the issue of cross-modal fusion bias in MAD, we systematically analyze three baseline models under both unimodal and multimodal settings using the trace of FIM. As shown in Figure \ref{rgbd_vis_1} (see Supplementary Materials for more results), we observe the models exhibit comparable information acquisition levels and similar variation trends when trained solely on RGB or depth modality. However, when RGB and depth are jointly utilized, all baselines consistently exhibit a clear imbalance in information acquisition (i.e., one modality is suppressed by the other, inducing cross-modal fusion bias) even though the suppressed modality performs well under unimodal training \cite{liu2025multi,zhang2023style}. Based on these observations, it is essential to further explore the potential impact of cross-modal fusion bias on MAD and elucidate its underlying mechanism, thereby developing an effective mitigation approach to break the performance bottleneck of existing methods.

\textbf{The Impact of Cross-modal Fusion Bias.} Naturally, one of the most direct effects of cross-modal fusion bias is suboptimal performance \cite{li2026cadtrack,li2026ragtrack}. To further explore the underlying mechanism of this bias that affects MAD, we carefully design five ablative studies. As shown in Figure \ref{fig1111_1_1_1} and Table \ref{tab1}, we observe that the fusion performance largely depends on the information acquisition of each modality in the early stages; interfering with the learning process in these stages will cause irreversible damage to the model’s capabilities. For example, perturbing the original input modality in early stages (i.e., blurring input images) consistently leads to significant performance degradation across all baselines, which cannot be remedied by simply extending the training duration. In contrast, similar interventions applied after this period have minimal impact. These results indicate that enhancing the information acquisition of suppressed modality during early stages, while preserving the information acquisition amount of dominant modality, can effectively prevent cross-modal fusion bias and improve fusion performance.

\section{Method}
\label{sec:Method}
In this section, we propose UCFB, a framework that can be flexibly embedded into existing MAD approaches to defy cross-modal fusion bias. Specifically, we outline the UCFB framework and elaborate on its learning algorithm.

\subsection{The Formulation of UCFB} The overview of UCFB is illustrated in Figure \ref{overall}. Unlike existing MAD paradigms that center on developing more sophisticated fusion strategies, UCFB is dedicated to mitigating cross-modal fusion bias and enhancing cross-modal interactions through unimodal adaptive adjustment (UAA) and canonical similarity analysis (CSA) modules. Specifically, inspired by \cite{peng2022balanced, wei2025fly}, unimodal adaptive adjustment dynamically monitors and calibrates modality-specific regularization weights using Fisher information during training. This adjustment is especially crucial during the critical learning epochs, as it prevents any single modality from acquiring information too rapidly and suppressing others. In addition, the CSA module encourages unimodal encoders to collaboratively capture shared and task-relevant information, thereby enhancing information acquisition in the suppressed modality. It is also worth highlighting that the UCFB framework is designed to facilitate integration with a variety of MAD methods, serving as a plug-and-play component. The following subsections detail their designs and learning algorithms.

\subsection{Characterizing Dominant Modality and Critical Learning Period} In UCFB, two key factors must be clarified before performing unimodal adaptive adjustment: (i) \textbf{how to identify the currently dominant modality}, and (ii) \textbf{how to determine critical learning periods}. For the former, suppose training is reached epoch $t$ and currently processing batch $b \in(0, B)$. The dominant modality is determined by computing the information acquisition gap ($\vartriangle_{\kappa}$) for each modality. Specifically, for modality $\kappa$ with the FIM trace $Tr(F_{\kappa;b}^t)$, we first calculate the number of other modalities whose FIM traces are smaller than $Tr(F_{\kappa;b}^t)$, denoted as $\mathcal{C}_\kappa$, as follows: 
\begin{equation}
\begin{split}
\vspace{-0.08in}
\mathcal{C}_\kappa=\sum_{\kappa' \ne \kappa; \,\, \kappa,\kappa' \in \mathbb{K}} \mathbb{I}\left[Tr(F_{\kappa ';b}^t) < Tr(F_{\kappa;b}^t)\right],
\end{split}
  \label{eq3_2_1}
\end{equation}
where $\mathbb{K}$ denotes the set of all modalities, and $\mathbb{I}[\cdot]$ is the indicator function. We then use $\vartriangle_{\kappa}$ to quantify the information acquisition gap between modality $\kappa$ and other modalities with lower or comparable information. A larger $\vartriangle_{\kappa}$ indicates stronger information acquisition ability of $\kappa$, as: 
\vspace{-0.08in}
\begin{equation}
\begin{split}
\vartriangle_{\kappa}=\frac{1}{\mathcal{C}_\kappa}\sum_{Tr(F_{\kappa ';b}^t) < Tr(F_{\kappa;b}^t)} \left[Tr(F_{\kappa ;b}^t) - Tr(F_{\kappa ';b}^t) \right],
\end{split}
  \label{eq3_2_2}
\end{equation}
Once the dominant modality $\kappa$ has been determined, we adopt Eq. \ref{eq3_2_3} to identify whether these modalities are in the critical period: 
\begin{equation}
\begin{split}
\frac{Tr(F_{\kappa;b}^{t-1})-Tr(F_{\kappa;b}^{t-2})}{Tr(F_{\kappa;b}^{t-1})}>\Upsilon,
\end{split}
  \label{eq3_2_3}
\end{equation}
where $\Upsilon$ is a threshold for determining whether to slow down the information acquisition of the dominant modality $\kappa$. If the value exceeds $\Upsilon$, the corresponding modality encoder is considered to be in a rapid information acquisition phase and is accordingly regulated to reduce its rate. 

\subsection{Unimodal Adaptive Adjustment} When slowing information acquisition in the dominant modality is necessary, the FIM trace, i.e., $Tr(F_{\kappa}^t)=Tr(F_{\kappa; 0}^t)+Tr(F_{\kappa; 1}^t)+...+Tr(F_{\kappa; B}^t)$ is obtained over the epoch $t$. The $Tr(F_{\kappa; b}^t)$ of batch $b$ is:
\begin{equation}
\begin{split}
Tr(F_{\kappa; b}^t)=\frac{1}{b}\sum_{i=0}^b \parallel g_{\psi^{\kappa}_i}^t\parallel^2,
\end{split}
  \label{eq3_3_1}
\end{equation}
where $g_{\psi^{\kappa}}$ denotes the gradient of modality $\kappa$. However, exactly computing $Tr(F_{\kappa}^t)$ requires access to the gradient information of the entire dataset, making it ineffective and impractical to dynamically adjust model information acquisition at each iteration of the $t$-th epoch via the full $Tr(F_{\kappa}^t)$. To address this issue, based on the gradient update rule (i.e., \fbox{$w_{\kappa;b}^t=w_{\kappa;b}^{t-1}-\xi\sum_{i=0}^bg_{\psi^{\kappa}_i}^t$}, where $\xi$ denote the learning rate), we approximately monitor the Fisher Information at each batch using a regularization term \fbox{$\Gamma_{\kappa,b}^t=\frac{\delta}{2}\parallel w_{\kappa;b}^t-w_{\kappa}^{t-1}\parallel^2$}, which can be written as:
\vspace{-0.1cm}
\begin{equation}
\begin{split}
&\Gamma_{\kappa,b}^t= \frac{\delta}{2}\parallel w_{\kappa;b}^t-w_{\kappa}^{t-1}\parallel^2
=\frac{\delta}{2}\parallel-\xi\sum_{i=0}^b g_{\psi^{\kappa}_i}^t\parallel^2,\\
&=\frac{\delta \xi^2}{2} \left\{\sum_{i=0}^{b} \parallel g_{\psi^{\kappa}_i}^t\parallel^2 +2\sum_{0\leq j \leq i \leq b} g_{\psi^{\kappa}_j}^t (g_{\psi^{\kappa}_i}^t)^T\right\},
\end{split}
  \label{eq3_3_2}
\end{equation}
Then, from \textbf{\textit{Lemma} 1}, the high dimensionality of $g_{\psi^{\kappa}_b}^t$ makes the gradients $g_{\psi^{\kappa}_j}^t (g_{\psi^{\kappa}_i}^t)^T$ of any two mini-batches $i$ and $j$ approximately orthogonal \cite{fisher1925theory, zhang2026gpr}. Thus, the $\Gamma_{\kappa,b}^t$ can be approximated as $\Gamma_{\kappa,b}^t=\frac{\delta \xi^2}{2}\sum_{i=0}^{b} \parallel g_{\psi^{\kappa}_i}^t\parallel^2$. Subsequently, we analyze the orthogonality between arbitrary batches $i$ and $j$:
\begin{commentbox}
    \textit{\textbf{Lemma}} \textbf{1.}\,\, In a high-dimensional space, let $g_{\psi^{\kappa}_i}^t$, $g_{\psi^{\kappa}_j}^t \in R^n$ denote two random vectors uniformly sampled from the surface of an $n$-dimensional hypersphere, with magnitudes $\parallel g_{\psi^{\kappa}_i}^t\parallel=\alpha$ and $\parallel g_{\psi^{\kappa}_j}^t\parallel=\beta$, respectively. As $n\rightarrow\infty$, $g_{\psi^{\kappa}_i}^t$ and $g_{\psi^{\kappa}_j}^t$ become approximately orthogonal; that is, their dot product satisfies:
\begin{equation}
\begin{split}
g_{\psi^{\kappa}_i}^t \cdot g_{\psi^{\kappa}_j}^t=\alpha \beta \cos \theta \approx0.
\end{split}
\label{pro3_3_3}
\end{equation}

\textbf{Proof.}\,\, Based on Eq.\ref{pro3_3_3}, to study the angular distribution ($\theta$) in high-
\end{commentbox}



\begin{commentbox}
dimensional spaces, we examine the geometry of the $n$-dimensional unit hypersphere. Any vector $\mathbb{V} \in R^n$ with unit norm ($||\mathbb{V}||_2=1$) resides on the surface of this hypersphere. Such a vector can be represented using n-dimensional spherical coordinates as: $\mathbb{V}=(v_1,v_2,v_3,...,v_n)$, where $v_i \in R^n$, $\sum_{i=1}^nv_i^2=1 $. When parameterized by spherical coordinates, $\mathbb{V}$ has the following component-wise form:
\begin{equation}
\begin{split}
v_1=\cos\varphi_1, \,\, v_2=\sin\varphi_1 &\cos\varphi_2, \,\,v_3=\sin\varphi_1 \sin\varphi_2 \cos\varphi_3,\\
... , v_{n-1}=(\sum_{i=1}^{n-2}sin\varphi_i) &\cos\varphi_{n-1},\,\, v_n=\sum_{i=1}^{n-1} \sin\varphi_i,
\end{split}
  \label{appendex1_2}
\end{equation}
where \fbox{$\varphi_1, \varphi_2, \varphi_3, ..., \varphi_{n-2} \in [0, \pi]$}, and \fbox{$\varphi_{n-1} \in [0, 2\pi]$}. Therefore, the hypersphere’s surface element ($d\rho$) can be expressed as:
\begin{equation}
\begin{split}
d\rho=&(\sin\varphi_1)^{n-2}(\sin\varphi_2)^{n-3}(\sin\varphi_3)^{n-4}...\\&\sin\varphi_{n-2}d\varphi_1d\varphi_2d\varphi_3...d\varphi_{n-1} .
\end{split}
  \label{appendex1_3}
\end{equation}
Without loss of generality, we align vector $g_{\psi^{\kappa}_i}^t$ with the $v_1$-axis, i.e., $g_{\psi^{\kappa}_i}^t=(\alpha,0,0,...,0)$. The second vector $g_{\psi^{\kappa}_j}^t$ is expressed in spherical coordinates, where the angle $\theta$ between $g_{\psi^{\kappa}_i}^t$ and $g_{\psi^{\kappa}_j}^t$ coincides with the first angular component $\varphi_1$, yielding:\fbox{$\cos\varphi_1=\cos\theta$}. The relevant term in the surface element of the hypersphere is:
\begin{equation}
\begin{split}
P_n(\varphi_1) \propto (\sin\varphi_1)^{n-2}
\end{split}
  \label{appendex1_3}
\end{equation}
This indicates that the probability density of $\varphi$ (or $\theta$) is proportional to the sine function raised to the power of $(n-2)$. When $n$ is large, the term $(\sin\varphi_1)^{n-2}$ becomes highly peaked around $\varphi_1 = \frac{\pi}{2}$, since sin $\varphi_1$ attains its maximum at that point. As $n$ approaches infinity, this peak sharpens further, making it very likely that $\varphi_1$ is close to $\frac{\pi}{2}$. Therefore, under this approximation, we have:
\begin{equation}
\begin{split}
\cos\varphi_1=\cos\theta\approx0.
\end{split}
  \label{appendex1_3}
\end{equation}
Therefore, in high-dimensional spaces, the angle $\theta$ between two randomly sampled vectors is predominantly concentrated around $\frac{\pi}{2}$, yielding: \fbox{$g_{\psi^{\kappa}_i}^t \cdot g_{\psi^{\kappa}_j}^t=\alpha \beta \cos \theta$}. This indicates that the vectors become approximately orthogonal in the limit as $n \to \infty$.

\textbf{\textit{\underline{Remark.}}} \,\, Although \textbf{Lemma 1} may stem from an idealized high-dimensional assumption, it serves as a theoretical analytical tool, effectively demonstrating that weakly correlated vectors are often approximately orthogonal in high-dimensional spaces, and has been widely adopted in previous studies \cite{fisher1925theory, peng2022balanced, DBLP:conf/aaai/WangCZCZ26}. In practice, even when this assumption does not strictly hold, the cross terms $\sum_{0 \leq j < i \leq b} g_{\psi^{\kappa}_j}^t (g_{\psi^{\kappa}_i}^t)^T$ are negligible, as gradients from different mini-batches typically exhibit low alignment.
\end{commentbox}

\begin{table*}[t!]
\begin{center}
\caption{I-AUROC scores on  MVTec 3D-AD. The \colorbox{mycolor1_1}{green} represents the results of baselines, and the \colorbox{mycolor1_2}{red} indicates our results.}
\setlength{\tabcolsep}{2.8mm} 
\renewcommand{\arraystretch}{1}
\adjustbox{width=0.98\linewidth}{
\begin{tabular}{c|l|c|c ccc ccc ccc |c}
\toprule
\multirow{1}{*}{} & \multirow{2}{*}{Method} & \multirow{2}{*}{Year} &\multirow{2}{*}{Bagel} &\multirow{2}{*}{Cable Gland} &\multirow{2}{*}{Carrot} &\multirow{2}{*}{Cookie}  &\multirow{2}{*}{Dowel} &\multirow{2}{*}{Foam} &\multirow{2}{*}{Peach} &\multirow{2}{*}{Potato} &\multirow{2}{*}{Rope} &\multirow{2}{*}{Tire} &\multirow{2}{*}{Mean} \\
& &&  &  &  &  &  &  &  &   &  &  \\
\bottomrule
\rowcolor{gray!15}\multicolumn{14}{c}{\textbf{ Single-class Setting (One model for one category)}}\\
\midrule[0.5pt] 
\multirow{15}{*}{\rotatebox{90}{RGB + Depth}}
&BTF &CVPR23 &91.8 & 74.8 & 96.7 & 88.3 &93.2 & 58.2 & 89.6 & 91.2 &92.1 & 88.6 & 86.5\\
&AST &WACV23 &98.3 & 87.3 &97.6 &97.1 &93.2 & 88.5 &97.4 &98.1 &100.0& 79.7 &93.7\\
&3D-ST &WACV23 &95.0 & 48.3 &98.6 &92.1 &90.5 & 63.2 &94.5&98.8 &97.6 & 54.2 & 83.3\\
&Shape\_Guided  &ICML23 &98.6 & 89.4 & 98.3 & 99.1 &97.6 & 85.7 &99.0 &96.5& 96.0 & 86.9 &94.7\\
&M3DM &CVPR23 &99.4 & 90.9 &97.2 &97.6 &96.0 &94.2 &97.3 &89.9 &97.2 & 85.0 & 94.5\\
&MMRD  &AAAI24&99.9 & 94.3 & 96.4 & 94.3 &99.2 &91.2 &94.9 &90.1 &99.4 &90.1 &95.0\\

\arrayrulecolor{lightgray}\cmidrule{2-14}\arrayrulecolor{black}
&\cellcolor{mycolor1_1}EasyNet  &\cellcolor{mycolor1_1}MM23 &\cellcolor{mycolor1_1}99.1 &\cellcolor{mycolor1_1}99.8 & \cellcolor{mycolor1_1}91.8 &\cellcolor{mycolor1_1}96.8 & \cellcolor{mycolor1_1}94.5&\cellcolor{mycolor1_1}94.5  & \cellcolor{mycolor1_1}90.5 &\cellcolor{mycolor1_1}80.7 &\cellcolor{mycolor1_1}99.4 & \cellcolor{mycolor1_1}79.3 & \cellcolor{mycolor1_1}92.6\\
&\cellcolor{mycolor1_2}EasyNet+UCFB &\cellcolor{mycolor1_2} - & \cellcolor{mycolor1_2}99.3 & \cellcolor{mycolor1_2}100.0 &\cellcolor{mycolor1_2} 93.1&\cellcolor{mycolor1_2} 95.9&\cellcolor{mycolor1_2} 95.2&\cellcolor{mycolor1_2}94.5&\cellcolor{mycolor1_2} 91.3&\cellcolor{mycolor1_2} 85.1&\cellcolor{mycolor1_2} 99.8&\cellcolor{mycolor1_2}83.8 &\cellcolor{mycolor1_2}93.8\\


\arrayrulecolor{lightgray}\cmidrule{2-14}\arrayrulecolor{black}
&\cellcolor{mycolor1_1}CFM & \cellcolor{mycolor1_1}CVPR24 & \cellcolor{mycolor1_1}99.4 & \cellcolor{mycolor1_1}88.8 & \cellcolor{mycolor1_1}98.4& \cellcolor{mycolor1_1}99.3& \cellcolor{mycolor1_1}98.0&\cellcolor{mycolor1_1}88.8& \cellcolor{mycolor1_1}94.1& \cellcolor{mycolor1_1}94.3& \cellcolor{mycolor1_1}98.0&\cellcolor{mycolor1_1}95.3 &\cellcolor{mycolor1_1}95.4\\
&\cellcolor{mycolor1_2}CFM+UCFB &\cellcolor{mycolor1_2} - & \cellcolor{mycolor1_2}99.6 & \cellcolor{mycolor1_2}90.8 &\cellcolor{mycolor1_2} 98.9&\cellcolor{mycolor1_2} 99.3&\cellcolor{mycolor1_2} 98.5&\cellcolor{mycolor1_2}90.1&\cellcolor{mycolor1_2} 94.4&\cellcolor{mycolor1_2} 95.2&\cellcolor{mycolor1_2} 98.7&\cellcolor{mycolor1_2}95.8 &\cellcolor{mycolor1_2}96.1\\

\arrayrulecolor{lightgray}\cmidrule{2-14}\arrayrulecolor{black}
& \cellcolor{mycolor1_1}IUF & \cellcolor{mycolor1_1}ECCV24 & \cellcolor{mycolor1_1}98.7 & \cellcolor{mycolor1_1}98.2 & \cellcolor{mycolor1_1}96.0& \cellcolor{mycolor1_1}91.3& \cellcolor{mycolor1_1}97.6& \cellcolor{mycolor1_1}94.1& \cellcolor{mycolor1_1}95.0&\cellcolor{mycolor1_1}92.2& \cellcolor{mycolor1_1}98.2&\cellcolor{mycolor1_1} 90.6&\cellcolor{mycolor1_1} 95.2\\
&\cellcolor{mycolor1_2}IUF+UCFB &\cellcolor{mycolor1_2} - & \cellcolor{mycolor1_2}100.0 & \cellcolor{mycolor1_2}100.0 &\cellcolor{mycolor1_2}98.8&\cellcolor{mycolor1_2} 93.7&\cellcolor{mycolor1_2} 98.9&\cellcolor{mycolor1_2}97.2&\cellcolor{mycolor1_2} 97.4&\cellcolor{mycolor1_2} 93.6&\cellcolor{mycolor1_2} 98.0&\cellcolor{mycolor1_2}92.9 &\cellcolor{mycolor1_2}97.1\\

\arrayrulecolor{lightgray}\cmidrule{2-14}\arrayrulecolor{black}
& \cellcolor{mycolor1_1}3D-ADNAS & \cellcolor{mycolor1_1}AAAI25 & \cellcolor{mycolor1_1}99.7 & \cellcolor{mycolor1_1}100.0 & \cellcolor{mycolor1_1}97.1& \cellcolor{mycolor1_1}98.6& \cellcolor{mycolor1_1}96.6& \cellcolor{mycolor1_1}94.8& \cellcolor{mycolor1_1}89.7&\cellcolor{mycolor1_1}87.3& \cellcolor{mycolor1_1}100.0& \cellcolor{mycolor1_1}86.7&\cellcolor{mycolor1_1} 95.1\\
&\cellcolor{mycolor1_2}3D-ADNAS+UCFB &\cellcolor{mycolor1_2} - & \cellcolor{mycolor1_2}100.0 & \cellcolor{mycolor1_2}99.8 &\cellcolor{mycolor1_2} 98.2&\cellcolor{mycolor1_2} 98.8&\cellcolor{mycolor1_2} 97.3&\cellcolor{mycolor1_2}95.4&\cellcolor{mycolor1_2} 91.6&\cellcolor{mycolor1_2} 89.3&\cellcolor{mycolor1_2} 100.0&\cellcolor{mycolor1_2}86.3 &\cellcolor{mycolor1_2}95.7\\
\bottomrule
\rowcolor{gray!15}\multicolumn{14}{c}{\textbf{ Multi-class Setting (One model for all categories)}}\\
\midrule[0.5pt] 
\multirow{12}{*}{\rotatebox{90}{RGB + Depth}} 
&M3DM & CVPR23 & 98.7 &84.7 &94.6& 95.6& 89.2&91.4& 92.3& 95.8& 91.6&87.3 &92.1\\
&DiAD  &AAAI24 &100.0 &73.9 &97.2 &71.6 &97.6 &98.7 &69.4  &78.3 &94.3 &85.6 &86.7\\
& MVFA & CVPR24 & 62.1 & 57.7 & 71.3& 73.7& 55.4& 53.6& 67.0&48.4& 91.6& 46.9& 62.7\\
& AdaCLIP & ECCV24 & 85.3 & 61.6 & 80.4& 86.4& 73.1&69.5& 77.4&63.3& 82.5& 61.6& 74.1\\
&CDAD  &CVPR25 &90.3 &89.7 &78.2 &68.4 &92.5 &73.4 &65.9 &63.3 &89.4 &79.7 &79.1\\
&AACLIP  &CVPR25 & 78.5 & 49.5 & 64.5& 99.1& 46.3&49.7& 87.6&87.5& 81.3&74.4 &71.8\\
&UniMMAD & - & 94.4 & 94.8 & 99.4& 94.2&95.9&89.1& 95.8& 93.3& 97.5&70.4 &92.5\\

\arrayrulecolor{lightgray}\cmidrule{2-14}\arrayrulecolor{black}
&\cellcolor{mycolor1_1}CFM & \cellcolor{mycolor1_1}CVPR24 & \cellcolor{mycolor1_1}99.2& \cellcolor{mycolor1_1}81.8 & \cellcolor{mycolor1_1}97.8& \cellcolor{mycolor1_1}98.9& \cellcolor{mycolor1_1}92.7&\cellcolor{mycolor1_1}91.4& \cellcolor{mycolor1_1}93.4& \cellcolor{mycolor1_1}88.4& \cellcolor{mycolor1_1}93.8&\cellcolor{mycolor1_1}86.6 &\cellcolor{mycolor1_1}92.4\\
&\cellcolor{mycolor1_2}CFM+UCFB &\cellcolor{mycolor1_2} - & \cellcolor{mycolor1_2}98.8 & \cellcolor{mycolor1_2}83.8 &\cellcolor{mycolor1_2} 98.0&\cellcolor{mycolor1_2} 98.5&\cellcolor{mycolor1_2} 93.6&\cellcolor{mycolor1_2}90.0&\cellcolor{mycolor1_2} 93.7&\cellcolor{mycolor1_2}90.3&\cellcolor{mycolor1_2} 94.4&\cellcolor{mycolor1_2}88.7 &\cellcolor{mycolor1_2}93.0\\

\arrayrulecolor{lightgray}\cmidrule{2-14}\arrayrulecolor{black}
&\cellcolor{mycolor1_1}IUF  &\cellcolor{mycolor1_1}ECCV24 &\cellcolor{mycolor1_1}97.2 &\cellcolor{mycolor1_1}96.9 &\cellcolor{mycolor1_1}89.1 &\cellcolor{mycolor1_1}75.4 &\cellcolor{mycolor1_1}93.4 &\cellcolor{mycolor1_1}89.7&\cellcolor{mycolor1_1}95.6 &\cellcolor{mycolor1_1}74.2 &\cellcolor{mycolor1_1}98.4 &\cellcolor{mycolor1_1}77.3 &\cellcolor{mycolor1_1}88.7\\
&\cellcolor{mycolor1_2}IUF+UCFB &\cellcolor{mycolor1_2} - & \cellcolor{mycolor1_2}98.0 & \cellcolor{mycolor1_2}97.3 &\cellcolor{mycolor1_2} 88.7&\cellcolor{mycolor1_2} 79.7&\cellcolor{mycolor1_2} 92.8&\cellcolor{mycolor1_2}90.4&\cellcolor{mycolor1_2} 96.8&\cellcolor{mycolor1_2}78.4&\cellcolor{mycolor1_2} 98.1&\cellcolor{mycolor1_2}78.5 &\cellcolor{mycolor1_2}89.9\\
\bottomrule
\end{tabular}
}
\end{center}
\vskip -0.1in
\label{tab2_mvt}
\end{table*} 
Following Eqs. \ref{eq3_3_2} and \ref{pro3_3_3}, $\Gamma_{\kappa,b}^t$ monitors $Tr(F_{\kappa; b}^t)$ to constrain the information acquisition of the dominant modality. To regulate the influence of $\Gamma_{\kappa,b}^t$, the parameter $\delta$ is defined as:
\vspace{-0.02in}
\begin{equation}
\begin{split}
\delta=exp(\mu * tanh(\vartriangle_{\kappa})),
\end{split}
  \label{eq3_3_4}
\end{equation}

where the hyperparameter $\mu$ controls the effect of $\delta$ on the information acquisition gap $\vartriangle_{\kappa}$, with larger $\delta$ yielding stronger regulation. Moreover, \textbf{\textit{Lemma}} \textbf{2} shows that, under appropriate parameter settings, introducing  $\Gamma_{\kappa,b}^t$ keeps the same convergence rate as the original objective.
\begin{commentbox}
    \textit{\textbf{Lemma}} \textbf{2.} At training epoch $t$ and batch $b$, let $\mathcal{L}$ denote the loss function, consider the optimization objective:
\begin{equation}
\begin{split}
\mathcal{L}(w_{\kappa ;b}^t)=\mathcal{L}_{total}(w_{\kappa ;b}^t)+\frac{\delta \xi^2}{2}\sum_{i=0}^{b} \parallel g_{\psi^{\kappa}_i}^t\parallel^2.
\end{split}
\label{pro3_3_5}
\end{equation} If $\delta$ and $\xi$ are sufficiently small, the convergence rate remains of the same order as without the $\Gamma_{\kappa,b}^t$.
\end{commentbox}

\textbf{Proof.} Due to the limitation of pages, the detailed proof is provided in the Supplementary Materials.

\subsection{Cross-modal Interaction Learning} While slowing information acquisition in the dominant modality, we introduce the canonical similarity analysis (CSA) module \cite{licsa, long2025enhancing} to encourage unimodal encoders to collaboratively extract shared and task-relevant information \cite{long2023deep,li2025frequency, zhao2026resilphase,TP-Seg,DifferSeg,xu2026hvpnet,liu2026opera}, thereby improving information acquisition in the suppressed modality. Specifically, let $\mathcal{M}_{r}$  and $\mathcal{M}_{d}$ denote the extracted RGB and Depth features, respectively. CSA then computes modality-specific representations from $\mathcal{M}_{r}$  and $\mathcal{M}_{d}$  via multiple nonlinear transformation layers \cite{li2026retrack,li2026conesep,li2026combiner}, denoted by $f_r$ and $f_d$:
\begin{equation}
\begin{split}
\mathbb{O}_{RGB}=f_{r} (\mathcal{M}_{r}; \mathcal{W}_{r}),
\mathbb{O}_{Depth}=f_{d} (\mathcal{M}_{d}; \mathcal{W}_{d}),
\end{split}
  \label{eq3_4_1}
\end{equation}
where $\mathcal{W}_{r}$ and $\mathcal{W}_{d}$ are the parameters of $f_r$ and $f_d$, respectively; $\mathbb{O}_{RGB}$ and $\mathbb{O}_{Depth}$ denote the output features processed by CSA modules. The objective is to jointly learn $\mathcal{W}_{r}$ and $\mathcal{W}_{d}$ by maximizing the correlation between $\mathbb{O}_{RGB}$ and $\mathbb{O}_{Depth}$, as defined below:
\begin{equation}
(\mathcal{W}_{r}^*;\mathcal{W}_{d}^*)=\mathop{\arg\max}\limits_{\mathcal{W}_{r}; \mathcal{W}_{d}} \,corr(f_{r}(\mathcal{M}_{r},\mathcal{W}_{r}),f_{d}(\mathcal{M}_{d},\mathcal{W}_{d})),
  \label{eq3_4_2}
\end{equation}
Subsequently, the features $\mathbb{O}_{RGB}$ and $\mathbb{O}_{Depth}$ are fused for subsequent anomaly discrimination. Building upon this, UCFB regulates information acquisition of dominant modality while enhancing cross-modal interaction and information sharing, thereby effectively mitigating cross-modal fusion bias and further improving overall MAD performance.

\begin{table*}[t!]
\begin{center}
\caption{I-AUROC scores on  Eyecandies. The \colorbox{mycolor1_1}{green} represents the results of baselines, and the \colorbox{mycolor1_2}{red} indicates our results.}
\setlength{\tabcolsep}{2.8mm} 
\renewcommand{\arraystretch}{1.1}
\adjustbox{width=0.98\linewidth}{
\begin{tabular}{c|l|c|c ccc ccc ccc |c}
\toprule
\multirow{1}{*}{} & \multirow{2}{*}{Method} & \multirow{2}{*}{Year} &\multirow{1}{*}{Candy} &\multirow{1}{*}{Chocolate}  &\multirow{1}{*}{Chocolate} &\multirow{2}{*}{Confetto} &\multirow{1}{*}{Gummy} &\multirow{1}{*}{Hazelnut}  &\multirow{1}{*}{Licorice}  &\multirow{2}{*}{Lollipop} &\multirow{1}{*}{Marsh-} &\multirow{1}{*}{Peppermint} &\multirow{2}{*}{Mean} \\
&& &Cane  &Cookie  &Praline  &  &Bear  &Truffle  &Sandwish  &   &mallow  &Candy  \\
\toprule
\rowcolor{gray!15}\multicolumn{14}{c}{\textbf{Single-class Setting (One model for one category)}}\\
\midrule[0.5pt] 
\multirow{13}{*}{\rotatebox{90}{RGB + Depth}}
&Voxel VM  &VISIGRAPP22 &55.3 & 77.2 & 48.4 & 70.1 &75.1 &57.8 & 48.0 &46.6 & 68.9 &61.1 &60.9\\
&AST &WACV23 &57.4 & 74.7 &74.7 &88.9 &59.6 & 61.7 &81.6 &84.1 &98.7& 98.7 &78.0\\
&M3DM &CVPR23 &62.4 & 95.8 & 95.8 &100.0 &88.6 & 78.5 &94.9 & 83.6 &100.0 & 100.0 & 89.7\\
&MMRD &AAAI24 &85.4 & 100.0 & 94.6 & 99.8 &90.8 &74.7 &96.6 &98.4 &100.0 &100.0 & 94.0\\

\arrayrulecolor{lightgray}\cmidrule{2-14}\arrayrulecolor{black}
&\cellcolor{mycolor1_1}EasyNet  &\cellcolor{mycolor1_1}MM23 &\cellcolor{mycolor1_1}73.7 &\cellcolor{mycolor1_1} 93.4 &\cellcolor{mycolor1_1}86.6 &\cellcolor{mycolor1_1}96.6 &\cellcolor{mycolor1_1}71.7 & \cellcolor{mycolor1_1}82.2 & \cellcolor{mycolor1_1}84.7 &\cellcolor{mycolor1_1}86.3 & \cellcolor{mycolor1_1}97.7 &\cellcolor{mycolor1_1} 96.0 & \cellcolor{mycolor1_1}86.9\\
&\cellcolor{mycolor1_2}EasyNet+UCFB &\cellcolor{mycolor1_2} - & \cellcolor{mycolor1_2}76.6 & \cellcolor{mycolor1_2}95.2 &\cellcolor{mycolor1_2} 87.3&\cellcolor{mycolor1_2} 96.8&\cellcolor{mycolor1_2} 73.4&\cellcolor{mycolor1_2}82.0&\cellcolor{mycolor1_2} 85.5&\cellcolor{mycolor1_2}88.2&\cellcolor{mycolor1_2}97.2&\cellcolor{mycolor1_2}95.7 &\cellcolor{mycolor1_2}87.8\\


\arrayrulecolor{lightgray}\cmidrule{2-14}\arrayrulecolor{black}
&\cellcolor{mycolor1_1}CFM &\cellcolor{mycolor1_1}CVPR24 &\cellcolor{mycolor1_1}68.0 & \cellcolor{mycolor1_1}93.1 & \cellcolor{mycolor1_1}95.2 & \cellcolor{mycolor1_1}88.0 &\cellcolor{mycolor1_1}86.5  &\cellcolor{mycolor1_1}78.2 &\cellcolor{mycolor1_1}91.7 &\cellcolor{mycolor1_1}84.0 &\cellcolor{mycolor1_1}99.8 &\cellcolor{mycolor1_1}96.2&\cellcolor{mycolor1_1}88.1\\
&\cellcolor{mycolor1_2}CFM+UCFB &\cellcolor{mycolor1_2} - & \cellcolor{mycolor1_2}69.8 & \cellcolor{mycolor1_2}94.2 &\cellcolor{mycolor1_2} 95.8&\cellcolor{mycolor1_2} 89.6&\cellcolor{mycolor1_2} 87.6&\cellcolor{mycolor1_2}79.1&\cellcolor{mycolor1_2} 93.0&\cellcolor{mycolor1_2} 86.4&\cellcolor{mycolor1_2} 99.6&\cellcolor{mycolor1_2}96.6 &\cellcolor{mycolor1_2}89.2\\

\arrayrulecolor{lightgray}\cmidrule{2-14}\arrayrulecolor{black}
& \cellcolor{mycolor1_1}IUF &\cellcolor{mycolor1_1} ECCV24 &\cellcolor{mycolor1_1} 61.6 &\cellcolor{mycolor1_1} 91.5 &\cellcolor{mycolor1_1} 82.8&\cellcolor{mycolor1_1} 95.4&\cellcolor{mycolor1_1} 75.2&\cellcolor{mycolor1_1} 66.7&\cellcolor{mycolor1_1} 84.7&\cellcolor{mycolor1_1}75.9&\cellcolor{mycolor1_1} 98.8&\cellcolor{mycolor1_1} 91.7&\cellcolor{mycolor1_1} 82.4\\
&\cellcolor{mycolor1_2}IUF+UCFB &\cellcolor{mycolor1_2} - & \cellcolor{mycolor1_2}62.3 & \cellcolor{mycolor1_2}92.3 &\cellcolor{mycolor1_2} 84.2&\cellcolor{mycolor1_2} 96.6&\cellcolor{mycolor1_2} 74.0&\cellcolor{mycolor1_2}67.2&\cellcolor{mycolor1_2} 86.2&\cellcolor{mycolor1_2} 78.5&\cellcolor{mycolor1_2} 99.4&\cellcolor{mycolor1_2}92.3 &\cellcolor{mycolor1_2}83.3\\

\arrayrulecolor{lightgray}\cmidrule{2-14}\arrayrulecolor{black}
& \cellcolor{mycolor1_1}3D-ADNAS & \cellcolor{mycolor1_1}AAAI25 &\cellcolor{mycolor1_1} 89.6 & \cellcolor{mycolor1_1}100.0 & \cellcolor{mycolor1_1}97.0& \cellcolor{mycolor1_1}100.0& \cellcolor{mycolor1_1}82.7& \cellcolor{mycolor1_1}88.2& \cellcolor{mycolor1_1}93.1&\cellcolor{mycolor1_1}95.0& \cellcolor{mycolor1_1}100.0& \cellcolor{mycolor1_1}100.0& \cellcolor{mycolor1_1}94.6\\
&\cellcolor{mycolor1_2}3D-ADNAS+UCFB &\cellcolor{mycolor1_2} - & \cellcolor{mycolor1_2}89.0 & \cellcolor{mycolor1_2}100.0 &\cellcolor{mycolor1_2} 97.4&\cellcolor{mycolor1_2} 99.6&\cellcolor{mycolor1_2} 84.3&\cellcolor{mycolor1_2}88.7&\cellcolor{mycolor1_2} 93.1&\cellcolor{mycolor1_2} 95.8&\cellcolor{mycolor1_2} 100.0&\cellcolor{mycolor1_2}99.8 &\cellcolor{mycolor1_2}94.8\\

\bottomrule
\rowcolor{gray!15}\multicolumn{14}{c}{\textbf{Multi-class Setting (One model for all categories)}}\\
\midrule[0.5pt] 
\multirow{11}{*}{\rotatebox{90}{RGB + Depth}} 
&M3DM & CVPR23 & 60.3 &74.4 &75.8& 89.1& 75.2&53.4& 79.8& 78.7& 91.7&95.3 &77.3\\
&DiAD  &AAAI24 &54.1 &87.5 &78.2 &91.7 &58.2 &55.7 &85.0 &71.9 &98.8 &88.3 &76.9\\
& MVFA & CVPR24 & 44.8 & 58.1 & 72.6& 66.6& 55.0& 60.0& 69.0&63.6& 82.1& 76.3& 64.8\\
& AdaCLIP & ECCV24 & 80.6 & 78.7 & 86.6& 87.8& 58.2& 37.6& 63.4&70.0& 82.3& 85.9& 73.1\\
&CDAD  &CVPR25 &66.9 &91.4 &75.3 &88.9 &74.2 &55.8 &86.1 &77.1 &94.0 &82.5 &79.2\\
&AACLIP  &CVPR25 & 37.8 &55.8 &54.2& 61.3& 52.2&42.6& 47.5& 42.5& 60.2&27.7 &48.1\\

\arrayrulecolor{lightgray}\cmidrule{2-14}\arrayrulecolor{black}
&\cellcolor{mycolor1_1}CFM & \cellcolor{mycolor1_1}CVPR24 &\cellcolor{mycolor1_1} 49.6 &\cellcolor{mycolor1_1}88.2 &\cellcolor{mycolor1_1}88.0& \cellcolor{mycolor1_1}88.2& \cellcolor{mycolor1_1}79.3&\cellcolor{mycolor1_1}71.7& \cellcolor{mycolor1_1}88.6&\cellcolor{mycolor1_1} 80.7& \cellcolor{mycolor1_1}97.8&\cellcolor{mycolor1_1}85.8 &\cellcolor{mycolor1_1}81.8\\
&\cellcolor{mycolor1_2}CFM+UCFB &\cellcolor{mycolor1_2} - & \cellcolor{mycolor1_2}51.6 & \cellcolor{mycolor1_2}88.2 &\cellcolor{mycolor1_2} 89.1&\cellcolor{mycolor1_2} 88.6&\cellcolor{mycolor1_2} 80.5&\cellcolor{mycolor1_2}71.7&\cellcolor{mycolor1_2} 88.9&\cellcolor{mycolor1_2}81.4&\cellcolor{mycolor1_2} 96.9&\cellcolor{mycolor1_2}86.9 &\cellcolor{mycolor1_2}82.4\\

\arrayrulecolor{lightgray}\cmidrule{2-14}\arrayrulecolor{black}
&\cellcolor{mycolor1_1}IUF  &\cellcolor{mycolor1_1}ECCV24 &\cellcolor{mycolor1_1}67.3 &\cellcolor{mycolor1_1}87.2 &\cellcolor{mycolor1_1}81.6 &\cellcolor{mycolor1_1}85.6 &\cellcolor{mycolor1_1}61.7 &\cellcolor{mycolor1_1}57.9 &\cellcolor{mycolor1_1}81.6 &\cellcolor{mycolor1_1}66.3 &\cellcolor{mycolor1_1}98.2 &\cellcolor{mycolor1_1}96.3 &\cellcolor{mycolor1_1}78.4\\
&\cellcolor{mycolor1_2}IUF+UCFB &\cellcolor{mycolor1_2} - & \cellcolor{mycolor1_2}66.5 & \cellcolor{mycolor1_2}89.1 &\cellcolor{mycolor1_2} 82.4&\cellcolor{mycolor1_2} 86.3&\cellcolor{mycolor1_2} 62.1&\cellcolor{mycolor1_2}58.5&\cellcolor{mycolor1_2} 82.2&\cellcolor{mycolor1_2} 68.9&\cellcolor{mycolor1_2} 98.6&\cellcolor{mycolor1_2}96.1 &\cellcolor{mycolor1_2}79.1\\
\bottomrule
\end{tabular}
}
\end{center}
\vskip -0.1in
\label{tab3}
\end{table*}
\section{Evaluation of the UCFB Framework}
\label{sec:Expriments}
We conduct extensive comparative experiments to validate the effectiveness of UCFB under single-, multi-class, and few-shot settings.
\subsection{Dataset and Experimental Setting}
\textbf{Dataset.} Following convention \cite{horwitz2023back, rudolph2023asymmetric, gu2024rethinking, miao2025robust}, we choose MVTec 3D-AD \cite{DBLP:conf/visapp/BergmannJSS22} and Eyecandies \cite{bonfiglioli2022eyecandies} datasets for evaluation. Both datasets consist of 10 categories, with MVTec 3D-AD sourced from real-world data acquisition and Eyecandies derived from virtual data synthesis.

\textbf{Setup.} To validate UCFB as a plug-in solution, we integrate it into four recent MAD methods—EasyNet \cite{chen2023easynet}, CFM \cite{costanzino2024multimodal}, IUF \cite{tang2024incremental}, and 3D-ADNAS \cite{long2025revisiting}—under single-class, multi-class, and few-shot settings. For fair comparison, we adopt the original experimental protocols of each method. For the implementation details and evaluation metrics, please refer to Section \ref{sec:Validation} and Supplementary Materials.

\begin{table}[t]
    \centering
    \caption{Few-shot test results on MVTec 3D-AD dataset.} \label{few_mvt}
    \renewcommand\arraystretch{1}
    \setlength{\tabcolsep}{5.0mm}{
    \resizebox{\linewidth}{!}{
    	{\begin{tabular}{l|ccccc}
    		\toprule[1.0pt]
    	\multirow{2}{*}{Method} &\multicolumn{4}{c}{MVTec 3D-AD (I-AUROC)} \\ 
    		&5-shot &10-shot &50-shot &Full \\ 
                \midrule[0.5pt]
BTF & 67.1 &69.5 &80.6 &86.5\\
AST & 68.0 &68.9 &79.4 &93.7\\
M3DM & 82.2 &84.5 &90.7 &94.5\\
\midrule[0.5pt]
\rowcolor{mycolor1_1} CFM &81.1 &84.5&90.6&95.4\\
\rowcolor{mycolor1_2} CFM+UCFB &81.4 &85.2 &91.8 &96.1\\
\midrule[0.5pt]
\rowcolor{mycolor1_1} 3D-ADNAS &82.6 &84.8 &89.0&95.1 \\
\rowcolor{mycolor1_2} 3D-ADNAS+UCFB &83.2 &85.3 &90.1 &95.7\\
            \bottomrule[1.0pt]
    \end{tabular}}}}
\end{table}

\begin{table}[t]
    \centering
    \caption{Few-shot test results on Eyecandies dataset.} \label{few_eye}
    \renewcommand\arraystretch{1}
    \setlength{\tabcolsep}{5.0mm}{
    \resizebox{\linewidth}{!}{
    	{\begin{tabular}{l|ccccc}
    		\toprule[1.0pt]
    	\multirow{2}{*}{Method} &\multicolumn{4}{c}{Eyecandies (I-AUROC)} \\ 
    		&5-shot &10-shot &50-shot &Full \\ 
                \midrule[0.5pt]
BTF & 65.2&68.5 &72.1 &74.0\\
AST & 63.3 &67.1 &73.9 &78.0\\
M3DM & 76.4 &82.4 &83.6 &88.2\\
\midrule[0.5pt]
\rowcolor{mycolor1_1} CFM &79.5 &83.8&85.2&88.1\\
\rowcolor{mycolor1_2} CFM+UCFB &80.3 &84.3 &86.1 &89.2\\
\midrule[0.5pt]
\rowcolor{mycolor1_1} 3D-ADNAS &77.5 &80.7 &86.8&94.6 \\
\rowcolor{mycolor1_2} 3D-ADNAS+UCFB &77.8 &81.5&87.4 &94.8 \\
            \bottomrule[1.0pt]
    \end{tabular}}}}
\end{table}
\subsection{Quantitative Evaluation}
\textbf{Comparison of Single- and Multi-class Settings.} Tables 2 and 3 report the performance comparison of UCFB integrated with four state-of-the-art MAD methods. The results demonstrate that incorporating UCFB consistently outperforms the corresponding baseline across different datasets as well as single-class and multi-class settings. In particular, adopting UCFB, I-AUROC achieves maximum gains of 1.9\% and 1.1\% on the two datasets under the single-class setting, and up to 1.2\% and 0.7\% under the multi-class setting, respectively, while maintaining comparable memory consumption and frame rates. Moreover, Figure~\ref{fig5_vis} further validates the effectiveness of the proposed approach, highlighting its advantages as a plug-and-play, MAD-friendly architecture. Overall, UCFB serves as a valuable complement to existing MAD methods by effectively alleviating cross-modal fusion bias and improving overall performance. Additional experimental results are provided in the Supplementary Materials.

\textbf{Comparison of Few-shot Settings.} To further scrutinize the effectiveness of UCFB, we report its experimental results under few-shot settings \cite{kim2024few, dai2025seas, lin2025commonality}. As displayed in Tables \ref{few_mvt} and \ref{few_eye}, we observe that UCFB achieves significant performance improvements over the baselines. Specifically, UCFB improves I-AUROC by 0.3\%, 0.7\%, and 1.2\% on MVTec 3D-AD when integrated into CFM, and by 0.3\%, 0.8\%, and 0.6\% on Eyecandies when integrated into 3D-ADNAS, under the 5-, 10-, and 50-shot settings. In this sense, these results indicate that mitigating cross-modal fusion bias from a Fisher information perspective is indeed an effective manner to advance MAD.

\subsection{Ablation Study}
\begin{figure}[t]
  \centering
  \setlength{\abovecaptionskip}{0.1cm}
      \includegraphics[scale=0.5]{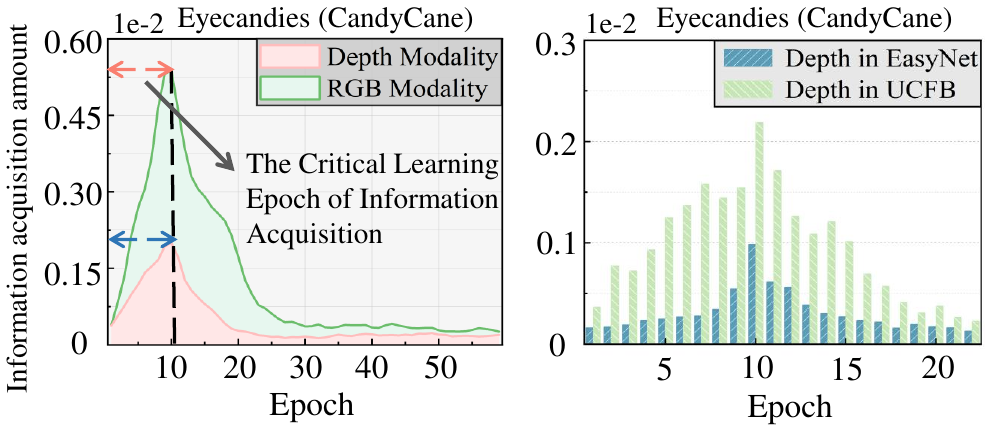}
  \caption{(\textbf{Left}) The value of FIM trace for the RGB and depth modalities in UCFB. (\textbf{Right}) The value of the FIM trace of the depth modality in UCFB compared with EasyNet.}
  \label{fig4_2_1}
\end{figure}

\begin{figure}[t]
  \centering
  \setlength{\abovecaptionskip}{0.1cm}
      \includegraphics[scale=0.37]{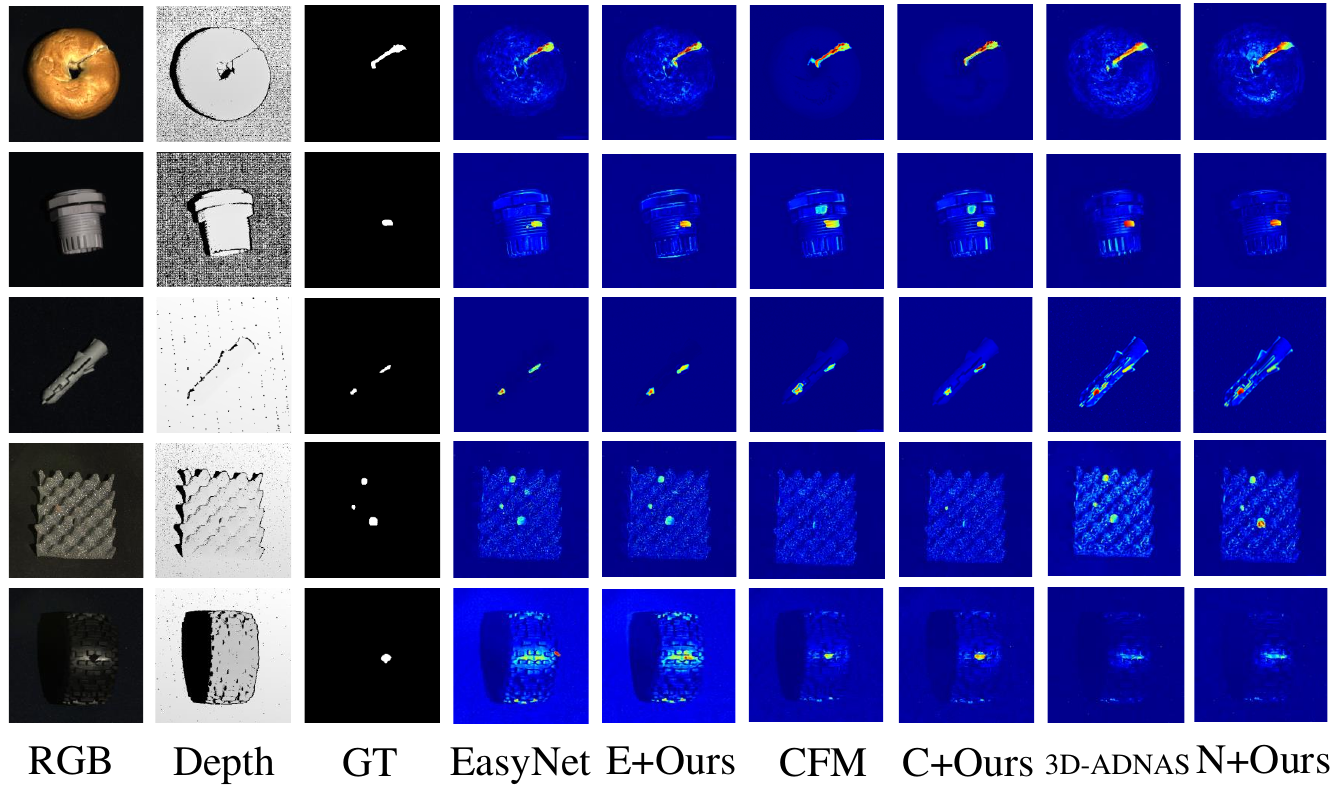}
  \caption{Visualization results between our method and EasyNet (E), CFM (C), and 3D-ADNAS (N) on the MVTec 3D-AD dataset.}
  \label{fig5_vis}
\vspace{-0.08in}
\end{figure}

\textbf{Evaluating Unimodal Adaptive Adjustment.} As shown in Figure \ref{fig4_2_1}-left, we first report the $Tr(F_{\kappa})$ values of the RGB and depth modalities when UCFB is integrated into EasyNet. The results show that $Tr(F_k)$ increases significantly for both modalities, implying that each modality acquires sufficient information in the early stages. Then, as shown in Figure \ref{fig4_2_1}-right, we further compare the $Tr(F_{\kappa})$ values of the depth modality with and without UCFB and observe that integrating UCFB consistently yields higher $Tr(F_{\kappa})$ values. This indicates that, during the critical learning epoch, UCFB enhances information acquisition of the suppressed modality while preserving that of the dominant modality, thereby effectively mitigating cross-modal fusion bias and improving overall model performance. 

\textbf{Impact of Key Components and Hyperparameters.} In this part, we again scrutinize the impact of key components in UCFB and their hyperparameter settings. As shown in Table 6, firstly, we observe that across all ablation tests, jointly incorporating canonical similarity analysis (CSA) and unimodal adaptive adjustment (UAA) modules within UCFB consistently yields superior performance improvements than using either module alone; for example, it outperforms the CSA-only and UAA-only variants by 1.8\% and 0.5\% in I-AUROC, respectively. In addition, we analyze the effects of the thresholds $\Upsilon$ and $\mu$ in UAA that regulate information acquisition of the dominant modality, and find that smaller values of $\Upsilon$ and larger $\mu$ lead to better performance.

\begin{table}[t]
\label{tab_abl_fin}
    \centering
    \caption{Ablation study on EasyNet using MVTec 3D-AD.} \label{Ablation study}
    \renewcommand\arraystretch{1}
    \setlength{\tabcolsep}{2.5mm}{
    \resizebox{\linewidth}{!}{
    	{\begin{tabular}{lc|cc|cc}
    		\toprule[1.0pt]
\rowcolor{gray!15}\multicolumn{2}{c|}{Ablation modules} &\multicolumn{2}{c|}{UAA parameter settings} &\multicolumn{2}{c}{Metric} \\ 
\rowcolor{gray!15}CSA &UAA & $\Upsilon$ & $\mu$ &I-AUROC &P-AUROC \\
\midrule[0.5pt]
\textcolor{forestgreen}{\XSolidBrush} &\textcolor{forestgreen}{\XSolidBrush}&None &None &92.6 &91.9 \\
\textcolor{red}{\textbf{\Checkmark}} &\textcolor{forestgreen}{\XSolidBrush}&None &None &92.6 &93.2 \\
\midrule[0.5pt]
\textcolor{forestgreen}{\XSolidBrush} &\textcolor{red}{\textbf{\Checkmark}}&0.04 &0.9 &92.8 &93.0 \\
\textcolor{forestgreen}{\XSolidBrush} &\textcolor{red}{\textbf{\Checkmark}}&0.01 &0.9 &93.3 &94.4\\
\textcolor{forestgreen}{\XSolidBrush} &\textcolor{red}{\textbf{\Checkmark}}&0.01 &0.5&92.9 &93.3  \\
\textcolor{forestgreen}{\XSolidBrush} &\textcolor{red}{\textbf{\Checkmark}}&0.01 &0.1&91.7 &92.1 \\
\midrule[0.5pt]
\textcolor{red}{\textbf{\Checkmark}} &\textcolor{red}{\textbf{\Checkmark}}&0.04 &0.9 &93.1 &93.5 \\
\rowcolor{gray!15}\textcolor{red}{\textbf{\Checkmark}} &\textcolor{red}{\textbf{\Checkmark}}&0.01 &0.9 &93.8 &94.6\\
\bottomrule[1.0pt]
\end{tabular}}}}
\end{table}
\section{Related Work}
\label{sec:Related_Work}
\textbf{Unsupervised Anomaly Detection.} Unsupervised anomaly detection (UAD) has attracted widespread attention in industrial intelligent manufacturing  \cite{strater2024generalad, cao2024adaclip, long2026towards}. Early research primarily focused on the development of 2D-UAD methods \cite{bergmann2019mvtec, jiang2022softpatch, sadikaj2025multiads} that rely solely on RGB images for flaw detection, and can be broadly divided into two categories: feature-embedding-based approaches \cite{defard2021padim, li2021cutpaste, rudolph2022fully, deng2022anomaly, liu2023simplenet}, which learn and match normal representations via memory banks \cite{mcintosh2024unsupervised, li2024target} or teacher–student models \cite{tien2023revisiting, gu2023remembering}, and reconstruction-based approaches \cite{you2022unified, zavrtanik2022dsr, schluter2022natural}, which identify anomalies through image reconstruction using autoencoders \cite{zavrtanik2021draem, zhang2024realnet} and diffusion models \cite{hu2024anomalydiffusion, he2024diffusion,li2025one, zhao2026seeing}. 

\textbf{Multimodal Anomaly Detection.} Recently, with the release of the MVTec 3D-AD \cite{DBLP:conf/visapp/BergmannJSS22} and Eyecandies \cite{bonfiglioli2022eyecandies} datasets, MAD has emerged as a promising direction \cite{zavrtanik2024cheating, liu2024learning, huang2024adapting, ma2025aa} in industrial inspection by fusing multi-source data, such as RGB and depth images, to improve detection performance \cite{chen2023easynet, tu2024self, li2024towards, li2025multi}. For example, M3DM \cite{wang2023multimodal} proposes a hybrid multimodal fusion strategy based on memory banks, yielding consistent performance improvements. CFM \cite{costanzino2024multimodal} proposes a framework that learns cross-modality feature mappings from nominal samples. 3D-ADNAS \cite{long2025revisiting} employs an architecture-oriented approach to develop category-specific fusion strategies. Other methods, including EasyNet \cite{chen2023easynet}, MMRD \cite{gu2024rethinking}, and FIND \cite{li2025find}, have further advanced MAD by leveraging diverse fusion schemes.

\textbf{Remark.} However, existing efforts rarely focus on the potential impact of cross-modality fusion bias on MAD. To bridge this gap, we systematically study the effect of this bias on detection performance through extensive experiments and propose a simple, efficient, and plug-in method to break the performance bottlenecks of existing approaches.

\section{Conclusion}\label{sec:Conclusion}
In this paper, we investigate the impact of cross-modal fusion bias on MAD tasks and propose UCFB, a novel architecture designed to address this challenge. Specifically, UCFB leverages unimodal adaptive adjustment and canonical similarity analysis modules to mitigate fusion bias and enhance cross-modal interaction, serving as a plug-and-play component for improving MAD. Extensive experiments across multiple datasets and various settings validate the effectiveness and superiority of our proposed method. We hope this work inspires further study on mitigating cross-modal fusion bias, thereby breaking the performance bottlenecks of existing MAD methods.

\section*{Acknowledgments}
This work is supported by the National Natural Science Foundation of China under Grant 62472079.

\bibliographystyle{ACM-Reference-Format}
\balance
\bibliography{sample-base}










\end{document}